\documentclass[runningheads]{llncs}

\usepackage{eccvabbrv}

\usepackage{graphicx}
\usepackage{booktabs}
\usepackage{multirow}
\usepackage{adjustbox}
\usepackage{makecell}
\usepackage{subcaption}

\usepackage[accsupp]{axessibility}  

\usepackage{orcidlink}

\begin{document}

\title{Spherical Linear Interpolation and Text-Anchoring for Zero-shot Composed \\ Image Retrieval} 

\titlerunning{Slerp + TAT}

\author{Young Kyun Jang\inst{1} \and
Dat Huynh\inst{1} \and
Ashish Shah\inst{1} \and \\
Wen-Kai Chen\inst{2} \and
Ser-Nam Lim\inst{2}}

\authorrunning{F.~Author et al.}

\institute{Meta AI \and University of Central Florida}

\maketitle

\begin{abstract}
Composed Image Retrieval (CIR) is a complex task that retrieves images using a query, which is configured with an image and a caption that describes desired modifications to that image. Supervised CIR approaches have shown strong performance, but their reliance on expensive manually-annotated datasets restricts their scalability and broader applicability. To address these issues, previous studies have proposed pseudo-word token-based Zero-Shot CIR (ZS-CIR) methods, which utilize a projection module to map images to word tokens. However, we conjecture that this approach has a downside: the projection module distorts the original image representation and confines the resulting composed embeddings to the text-side. In order to resolve this, we introduce a novel ZS-CIR method that uses Spherical Linear Interpolation (Slerp) to directly merge image and text representations by identifying an intermediate embedding of both. Furthermore, we introduce Text-Anchored-Tuning (TAT), a method that fine-tunes the image encoder while keeping the text encoder fixed. TAT closes the modality gap between images and text, making the Slerp process much more effective. Notably, the TAT method is not only efficient in terms of the scale of the training dataset and training time, but it also serves as an excellent initial checkpoint for training supervised CIR models, thereby highlighting its wider potential. The integration of the Slerp-based ZS-CIR with a TAT-tuned model enables our approach to deliver state-of-the-art retrieval performance across CIR benchmarks.

\end{abstract}

\section{Introduction}
\label{sec:Introduction}

Composed Image Retrieval (CIR) is a vision-language task that utilizes both image and text queries to retrieve images with high precision. It aims to identify images that are visually similar to a reference image, but also incorporate changes specified in a text query, thereby providing users with precise control over the characteristics of the desired image. This bi-modality of the query allows for a more nuanced search, as some features are better described with language, while others are more effectively expressed visually. Due to its potential in a variety of real-world applications, there has been a surge in attention towards CIR.

Previous works on supervised CIR \cite{Combiner,ARTEMIS,CIRPLANT,FashionVLP} have proposed solutions to various problems using natural image \cite{CIRR,CASE,SEARLE} and fashion image \cite{fashionIQ} datasets, configured with triplets of <reference image, textual intent, target image>. However, these solutions face significant challenges, primarily due to the complexity and cost of dataset collection. The process involves creating these triplets, which is not easily automated and requires substantial manual labor. This makes the process time-consuming and resource-intensive, particularly when creating large training sets. Moreover, models trained on these supervised datasets tend to be specialized to specific use-cases, limiting their robustness to diverse unseen domains. Therefore, despite promising results from supervised CIR approaches, their reliance on expensive manually-annotated datasets restricts their scalability and broader applicability.

\begin{figure}[!t]
\centering
  \subcaptionbox{Pseudo-word token-based ZS-CIR method.
  \label{fig:pseudo_word_based}}{\includegraphics[width=9cm]{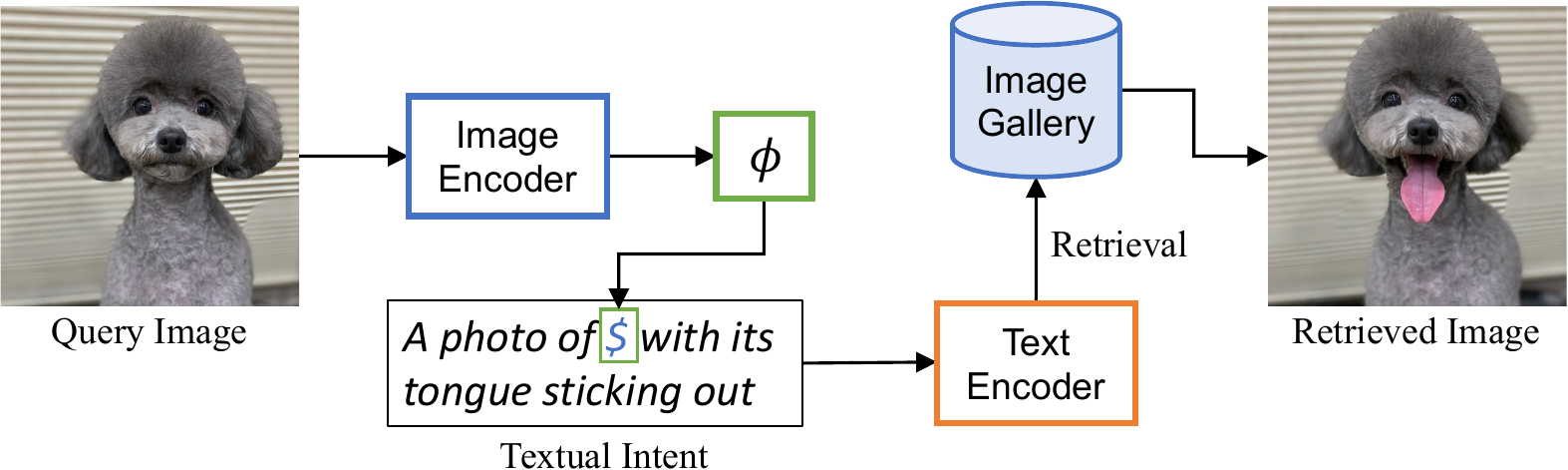}}
  \subcaptionbox{Our linear interpolation based ZS-CIR method.
  \label{fig:Slerp_based}}{\includegraphics[width=9cm]{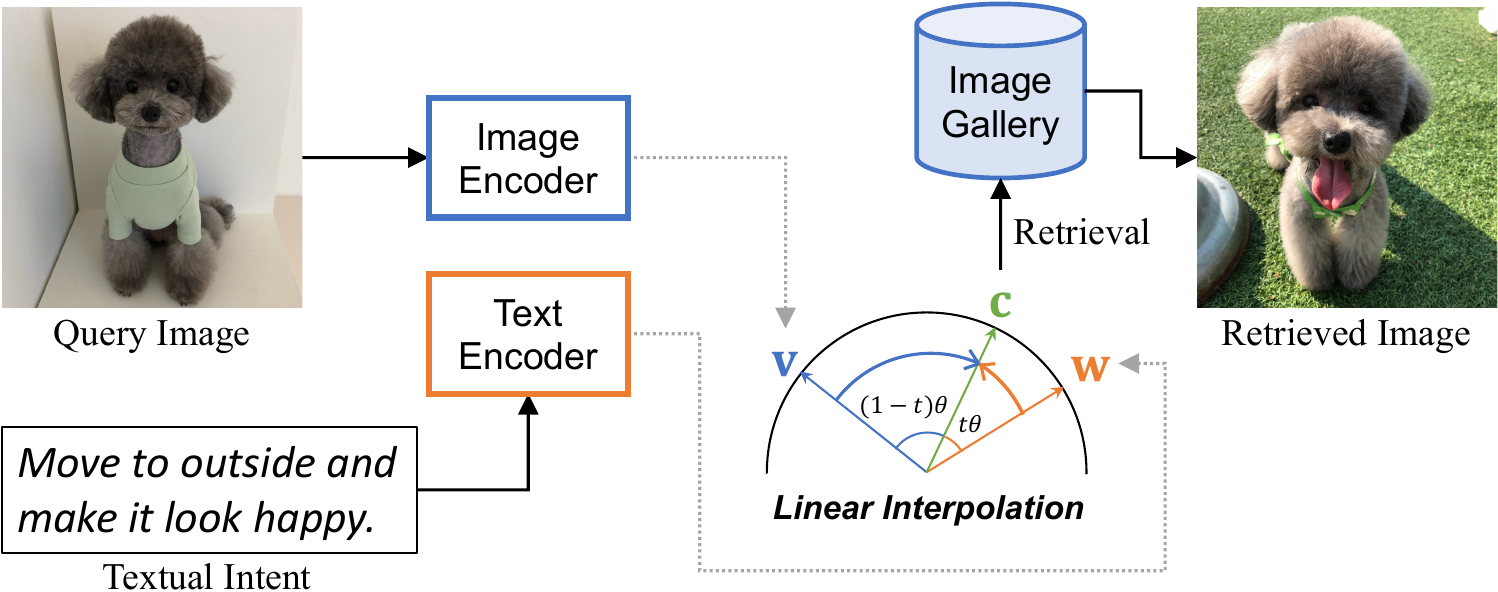}}
\caption{Overview of ZS-CIR approaches. The previous works \cite{Pic2word,PALAVRA,SEARLE} utilize a projection module, which transforms an image into a textual pseudo-word. This is then combined with text (textual intent) to produce a composed embedding with a text encoder for retrieval purposes. In contrast, we propose a method based on a simple spherical linear interpolation. This method directly combines image (\textbf{v}) and text (\textbf{w}) embeddings to produce a composed embedding (\textbf{c}). We then use \textbf{c} to perform ZS-CIR.}
\label{fig:introduction}
\end{figure}

In response to the challenges faced in supervised CIR, the research community has introduced an alternative approach known as Zero-Shot CIR (ZS-CIR) \cite{Pic2word,PALAVRA,SEARLE,CoVR,CASE}. ZS-CIR, unlike supervised CIR that relies on triplets, utilizes <image, text caption> pairs, akin to those used in the Vision-Language Pretraining (VLP) models like CLIP \cite{CLIP} or BLIP \cite{BLIP}. The primary challenge in ZS-CIR is establishing a compositional understanding between image and text. To address this, pseudo-word token-based approaches \cite{Pic2word,PALAVRA,SEARLE,LinCIR} were proposed as shown in Figure \ref{fig:pseudo_word_based}, aiming to utilize a projection module that transforms VLP image encoder output as the pseudo-word token. Then, this pseudo-word token is concatenated with text tokens and forward to VLP text encoder to produce image-text composed representation embedding to perform ZS-CIR. 

However, pseudo-word token-based methods may not genuinely "compose image and text" since the representation of the composed embeddings is constrained within the text encoder output embedding space, which may not fully capture the joint image-text information. Moreover, these methods have a limitation in their inference pipeline because the projection module can distort the original, discriminative image representation. This results in a CIR model that is less capable of capturing and representing the complex compositional relationships between image and text samples.

To avoid these problems, we propose a new method called \textbf{S}pherical \textbf{l}inear Int\textbf{erp}olation (Slerp \cite{Slerp})-based Zero-shot CIR. Given that VLP encoders are trained with scaled cosine similarity between image and text embeddings, this results in the distribution of image and text samples on a joint hypersphere with the radius of the scaling factor (temperature). Consequently, Slerp can be applied to find an intermediate embedding of image and text ones, as shown in Figure \ref{fig:Slerp_based}. Surprisingly, the results from Slerp-based retrieval confirm that simple interpolation between image and text embeddings of VLP can achieve ZS-CIR performance comparable to the best-performing approaches, without any additional training.

Furthermore, we introduce a \textbf{T}ext-\textbf{A}nchored-\textbf{T}uning (TAT) strategy to facilitate the Slerp process by reducing the gap between modalities. Based on the observations in \cite{Pic2word,SEARLE}, which show that the text-only case significantly outperforms the image-only case in CIR benchmarks and sometimes even surpasses the best performing model, we believe that the VLP text embedding itself plays a critical role in CIR. Therefore, we keep the VLP text encoder frozen to maintain its power and allow the text embeddings to serve as an \textit{anchor} for contrastive learning. Specifically, we apply LoRA \cite{LoRA} parameters to the VLP image encoder, which not only preserves the original knowledge of the image encoder but also effectively redistributes image embeddings to align more closely with the corresponding text embeddings. As a result, the modality gap is reduced while the text representations of VLP are retained. Ultimately, by combining Slerp and TAT, we significantly boost the performance of ZS-CIR.

Extensive experimental results on various domain benchmarks, including natural images \cite{CIRR,SEARLE} and fashion images \cite{fashionIQ}, underscore the advantages of Slerp-based ZS-CIR. This not only achieves decent retrieval performance, but Slerp also allows for easy adjustment of the contribution of image or text queries based on the user's intent. Our TAT strategy significantly enhances retrieval performance, even when trained with far fewer image-text training pairs (only 20\% of the number used in the CC3M dataset) compared to previous ZS-CIR methods \cite{Pic2word,LinCIR,CoVR,CASE}. Moreover, optimal performance of Slerp + TAT method can be achieved with just a single epoch of training, exhibiting superior training efficiency compared to pseudo-word token-based approaches \cite{Pic2word,PALAVRA,SEARLE,LinCIR}, which require tens to hundreds of training epochs. As part of the broader applications of our work, we demonstrate that TAT-trained VLP models can serve as superior initial checkpoints compared to the original VLP models when training supervised CIR models.

\section{Related Work}
\label{sec:Related Work}

\subsection{Supervised Composed Image Retrieval}
The field of image retrieval, as surveyed in \cite{IR_survey}, has attracted significant attention from researchers due to its diverse utilities, such as enhancing search engines and e-commerce platforms. In particular, Composed Image Retrieval (CIR), a method that retrieves images using a pair consisting of a query image and a text that depicts user intent \cite{Composing}, has been widely explored. This expansion from vision-only to vision-language multi-modal applications represents a significant advancement in the field. CIR has been explored in various visual domains, including natural \cite{CIRR,Naturalist,SEARLE} and fashion \cite{Dialog,Auto_spatially,FashionVLP,fashionIQ} images. Notably, CIRR \cite{CIRR} and FashionIQ \cite{fashionIQ} are widely utilized to train CIR models in a supervised fashion, using human-annotated triplets that contains a reference image, a target image, and a textual description of their difference. Based on this supervision, various supervised CIR methods \cite{ARTEMIS,CIRPLANT,Combiner,CIRR,FashionVLP} have been proposed, and they have demonstrated decent CIR performances. However, these methods often encounter bottlenecks due to the high cost of labeling and the potential existence of mislabeled samples. Additionally, the current supervised CIR datasets are limited to specific image domains and visual attributes, which restrict their applicability across different domains.

\subsection{Zero-shot Composed Image Retrieval}
As an alternative to supervised CIR, Zero-Shot CIR (ZS-CIR) methods \cite{Pic2word,SEARLE,PALAVRA,LinCIR} have been introduced. Instead of using CIR triplets, these zero-shot methods employ image-text caption pairs, similar to those used in the pretraining stage of VLP models \cite{CLIP,BLIP}. ZS-CIR is particularly advantageous for domain generalization as it can utilize simple web-crawled data samples for training, thereby making it easy to scale up the dataset size and incorporate a variety of image domains. This, in turn, can be used to build a robust CIR model. The primary challenge in ZS-CIR is how to achieve composition between image and text. To address this, previous works \cite{Pic2word,SEARLE,PALAVRA,LinCIR} have used pseudo-word token-based methods, which replace specific characters in text prompts with pseudo-word tokens derived from images. However, this approach does have its limitations. The image representation, which is inherent in the embedding, can become distorted when it is transformed into a pseudo-word token. Furthermore, this token is then processed through a fixed text encoder. This encoder has only been trained with text inputs, and as such, it may not correctly interpret the information from images. To overcome these issues, we propose a novel ZS-CIR scheme based on Slerp, which simply interpolates between image and text embeddings to obtain a composed representation for retrieval. In this manner, the image embedding directly contributes to building the composed embedding close to its original form. Moreover, we further improve the Slerp-based search with text-anchored fine-tuning to reduce the domain gap in VLP embeddings, thereby facilitating the Slerp process. It's important to note that our TAT differs from the training scheme in \cite{LiT}, which also proposed to freeze a single-modality tower (the image encoder) during the pretraining stage of the VLP model. In contrast, TAT fixes the text encoder and is focused on fine-tuning the existing VLP model to enhance its performance for CIR tasks.

\section{Method}
\label{sec:Method}

In this section, we present our Slerp-based ZS-CIR approach. Then, we introduce the TAT learning method, which aligns image representation closer to a fixed text, thereby enhancing the Slerp-based ZS-CIR performance significantly. We employ the pre-trained vision-language models, CLIP \cite{CLIP} and BLIP \cite{BLIP}, as our baseline models. Each of these models is equipped with a separate image and text encoder that generate visual and textual representation embeddings, respectively. During the retrieval stage, the composed representation embedding of the image and text query is obtained using Slerp. This composed embedding is then used to compute the similarity with pre-computed gallery image embeddings, resulting in ranked results.

\subsection{Preliminaries}
\label{sec:Preliminaries}

Represented by CLIP \cite{CLIP} and BLIP \cite{BLIP}, VLP models are trained on an extensive dataset $\mathcal{D}=\{(x_n, t_n)\}_{n=1}^N$ of image-text caption pairs, where $x_n$ is an image and $t_n$ is a tokenized text caption. The VLP model employs a parameterized image encoder $E_{I}$ and text encoder $E_{T}$ to produce an image embedding $\textbf{v}\in \mathbb{R}^d:\textbf{v}=E_{I}(x)$, and a text embedding $\textbf{w}\in \mathbb{R}^d:\textbf{w}=E_{T}(t)$ of the same dimensionality $d$, respectively. Both the image and text embeddings are \textit{l2}-normalized to utilize cosine similarity as a baseline metric.

Then, normalized temperature-scaled cross entropy loss \cite{SimCLR} (contrastive loss) is utilized to update trainable parameters of both modality encoders as:

\begin{equation}
\mathcal{L}_{cont.} = \mathcal{L}_{I2T} + \mathcal{L}_{T2I} 
 \label{eqn:contrastive_loss}
\end{equation}

\noindent where each term is described as:

\begin{align}
\mathcal{L}_{I2T}(\mathcal{B}) &= -\frac{1}{N_B}\sum_{i=1}^{N_B}\log\frac{\exp{(\textbf{v}_i^T\cdot\textbf{w}_i/\tau)}}{\sum_{j=1}^{N_B}\exp{(\textbf{v}_i^T\cdot\textbf{w}_j/\tau})}, \\
\mathcal{L}_{T2I}(\mathcal{B})
&= -\frac{1}{N_B}\sum_{i=1}^{N_B}\log\frac{\exp{(\textbf{w}_i^T\cdot\textbf{v}_i/\tau)}}{\sum_{j=1}^{N_B}\exp{(\textbf{w}_i^T\cdot\textbf{v}_j/\tau)}}.
\end{align}

\noindent Here, $\mathcal{B}$ denotes a training batch consisting of $N_B$ image-text pairs, and $\tau$ is the temperature parameter used for scaling. This learning strategy results in an image and its corresponding text caption being aligned, while the unpaired ones are separated.

\subsection{Spherical Linear Interpolation-based Retrieval}
\label{sec:Spherical Linear Interpolation-based Retrieval}

Due to the cosine similarity-based representation learning in the VLP model, both the image and text embeddings are distributed on a unique hypersphere with a radius of the scaling factor, $\tau$. Moreover, since the VLP model is trained on hundreds of millions of image-text pairs, the embeddings of images and texts are densely distributed on this hypersphere, presumably covering most of the related semantics between the vision and language domains.

Based on this, we assume that intermediate representation of existing image and text embeddings ($\textbf{v}$ and $\textbf{w}$) can represent image and text composed representation, and introduce a simple Slerp \cite{Slerp}-based approach for ZS-CIR as:

\begin{equation}
\textbf{c}:\text{Slerp}(\textbf{v}, \textbf{w}; \alpha) = \frac{\sin((1-\alpha)\theta)}{\sin(\theta)}\textbf{v} + \frac{\sin(\alpha\theta)}{\sin(\theta)}\textbf{w}
 \label{eqn:slerp}
\end{equation}

\noindent where $\theta$ is the angle between $\textbf{v}$ and $\textbf{w}$ which is obtained as:

\begin{equation}
\theta = \cos^{-1}(\textbf{v} \cdot \textbf{w})
\label{eqn:theta}
\end{equation}

\noindent and $\alpha$ is a balancing scalar value within the range of $[0, 1]$ that determines the scale of the linear combination between $\textbf{v}$ and $\textbf{w}$ to produce the composed embedding: $\textbf{c}$. It's worth noting that with the Slerp-based ZS-CIR method, users have the ability to manually balance the contributions of the query image or text by adjusting the value of $\alpha$ to suit their retrieval purposes.

In line with the experimental results reported in previous works, such as Pic2Word \cite{Pic2word} and SEARLE \cite{SEARLE}, we also observe a similar phenomenon where using a text-only query yields better ZS-CIR performance than an image-only query for both natural and fashion image dataset benchmarks. This observation leads us to hypothesize that text plays a more significant role than images in constructing the composed representation. As a result, we assign more weight ($\alpha \geq 0.8$; \textit{text-weighted}) to $\textbf{w}$ than to $\textbf{v}$ when constructing $\textbf{c}$, which leads to a significant performance gain in ZS-CIR evaluation protocols. That said, as we will show in Section \ref{sec:Experiments}, the image still plays an important role in Slerp. When the text is given all the weight, we observed significant performance degradation.

\subsection{Text-Anchored-Tuning}
\label{sec:Text-Anchored-Tuning}

\begin{figure}[!t]
\centering
\includegraphics[width=0.6\linewidth]{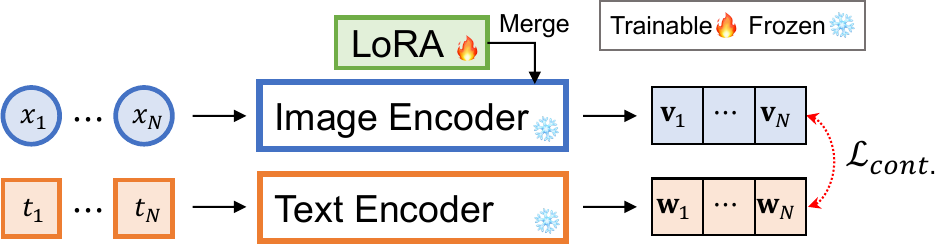}
\caption{Workflow of Text-Anchored-Tuning.}
\label{fig:TAT}
\vspace{-1em}
\end{figure}

In reality, the vision (image) and language (text) modality embeddings of the VLP model are distinctly distributed, and the embeddings of paired image and text samples are actually not closely aligned with each other as observed in \cite{Mindthegap}. This gap between modalities can limit the effectiveness of conducting interpolation between image and text embeddings, which may consequently degrade the Slerp-based ZS-CIR performance. To resolve this issue, we propose Text-Anchored-Tuning (TAT) method as illustrated in Figure \ref{fig:TAT}.

The goal of TAT is to mitigate the performance degradation in the text-weighted Slerp-based search, which is caused by the modality gap, while preserving the powerful original text representation of the VLP model. We aim to achieve this by adjusting the image embeddings to align with the text embedding points that we keep fixed (\textit{text-anchoring}), thereby making images closer to the corresponding texts. This is enabled by fine-tuning the image encoder $E_{I}$, while keeping the text encoder $E_{T}$ unchanged. However, fully fine-tuning all parameters of $E_{I}$ is not only expensive but can also lead to catastrophic forgetting of the inherent knowledge within $E_{I}$. Therefore, we adopt the LoRA technique \cite{LoRA} with additional trainable parameters ($\mathcal{P}_{lora}$), which retains the parameters of $E_{I}$ and adds a small number of additional parameters to $E_{I}$ to assist in its adaptation. In this way, the original expressive power of both $E_{I}$ and $E_{T}$ is preserved, and the image embeddings can be realigned to follow the text embeddings. For the training objective, we choose the same batch-wise contrastive loss (Equation \ref{eqn:contrastive_loss}) as used in the pretraining stage of VLP, with a fixed temperature ($\tau$) to ensure stable training. In the end, image embeddings follow the text representation to ease the text-weighted Slerp process.

\subsection{Inference}
\label{sec:Inference}

To perform composed image retrieval, we first construct a gallery. This is done by forwarding images ($x_g$) through the trained image encoder ($E_{I}$) to produce image embeddings ($\textbf{v}_g=E_{I}(x_g)$), which are then collected to form the retrieval gallery. Next, we take a user's query of an image ($x_q$) and text ($t_q$) and forward them into $E_{I}$ and the text encoder ($E_{T}$) separately, to generate the query image embedding ($\textbf{v}_q=E_{I}(x_q)$) and text embedding ($\textbf{w}_q=E_{T}(t_q)$). The image and text embeddings, $\textbf{v}_q$ and $\textbf{w}_q$, are then combined using Slerp with a balancing scale ($\alpha$) to produce a query composed embedding $\textbf{c}_q$. Finally, $\textbf{c}_q$ is used to retrieve relevant images from the gallery by computing the cosine similarity between $\textbf{c}_q$ and $\textbf{v}_g$, producing a ranked list based on the score. Note that, unlike pseudo-token based approaches \cite{Pic2word,PALAVRA,SEARLE,LinCIR}, our Slerp-based method does not require a specific text prompt (\eg "a photo of [\$]", where \$ is replaced with a text token transformed from an image). This requirement can often become a bottleneck for the performance of ZS-CIR. Instead, our Slerp-based search allows the user's query text to be directly applied, providing a more robust solution.

\section{Experiments}
\label{sec:Experiments}

\subsection{Settings}
\label{sec:Settings}

\noindent \textbf{Datasets.} For training ZS-CIR models, previous works \cite{Pic2word,LinCIR} have used datasets configured with web-collected image-text pairs such as CC3M \cite{CC3M}. We adopt three training datasets: (1) CC3M: we attempted to utilize the full scale of the CC3M dataset; however, we were only able to download 2.3M, so we use a subset of the entire package, (2) LLaVA-Align: this dataset contains image-text pairs used in LLaVA \cite{LLaVA} to build alignment between image and text modalities, which contains 585K pairs, and (3) Laion-2M: we sub-sample 2M of the Laion-115M image-text pairs used in the pretraining of the BLIP \cite{BLIP} model. Unless otherwise stated, we use this dataset as our baseline for the CIR model training.

For evaluation, there are three standard benchmarks used by ZS-CIR methods \cite{Pic2word,SEARLE,LinCIR}. The first is CIRR \cite{CIRR}, which deals with natural images. The second is CIRCO \cite{SEARLE}, which handles more challenging natural image cases on a larger scale than CIRR and is specifically designed for ZS-CIR models. The third is FashionIQ \cite{fashionIQ}, which focuses on fashion domain images. Each of these benchmarks offers unique challenges and datasets, aiding researchers in expanding the limits of what can be achieved in CIR. Specifically, \textit{CIRR} is configured with 4,351 subgroups (subsets), each containing six similar images, sourced from NLVR2 dataset \cite{NLVR2}. We utilize the test split of CIRR, which consists of 503 subgroups of 2,178 images, to evaluate ZS-CIR methods. \textit{CIRCO} contains 800 image-text queries collected from the COCO 2017 unlabeled set \cite{COCO}, with a retrieval gallery size of 123,403 for evaluation. \textit{FashionIQ} is divided into three categories: Dress, Shirt, and Toptee. For this benchmark, we employ the validation split, which consists of triplets of 15,415 images.

\noindent \textbf{Evaluation Metrics.} Following the protocols utilized in benchmarks \cite{CIRR,fashionIQ,SEARLE}, we report the CIR results with recall scores at the top K retrieval results (R@K) for CIRR and FashionIQ, and more detailed results under the collected subset ($\text{R}_{s}$@K) for CIRR. In the case of CIRCO, we follow the protocol used in \cite{SEARLE,LinCIR} and evaluate retrieval results with a ranking-based metric, mean Average Precision (mAP@K) scores.

\noindent \textbf{Implementation Details.} The popular CLIP and BLIP models \cite{CLIP,BLIP}, which are based on a Transformer \cite{Transformer} backbone, serve as our baseline VLP models. These models are listed with backbone/image patch size as: CLIP-ViT-B/32, CLIP-ViT-L/14, and BLIP-ViT-L/16. For simplicity, we notate each as C-B32, C-L14 and B-L16, respectively. Pretrained weights provided by HuggingFace\footnote{https://huggingface.co/models} \cite{Huggingface} are applied to the CLIP models as: \texttt{openai/clip-vit-base-patch32}, \texttt{openai/clip-vit-large-patch14}. For the BLIP model, the original checkpoint named \texttt{BLIP w/ ViT-L} from the authors\footnote{https://github.com/salesforce/BLIP} is employed.

For TAT learning (refer to Section \ref{sec:Text-Anchored-Tuning}), we apply additional LoRA parameters ($\mathcal{P}_{lora}$) configured as follows: $\text{LoRA}_\alpha=16$, $\text{rank}=16$, and $\text{dropout}=0.1$. We train the models using eight A100-80GB GPUs. The batch size is set to 1,024, with 128 batches per GPU. During training, we keep the entire set of parameters of $E_I$ and $E_T$ fixed. The temperature hyper-parameter $\tau$ is set to 1/0.07. We use the AdamW optimizer \cite{AdamW} with a fixed learning rate of 1e-4 and a weight decay of 0.01. The training, which is conducted in a \textit{single epoch} and where the trainable parameters constitute \textit{less than 0.5\% of the total parameters}, is performed in a highly efficient manner, \eg, taking less than 0.5 hour for C-B32 backbone with LLaVA-Align dataset.

During the inference stage, we observe that the CIRR is more heavily influenced by the text than the image, compared to other datasets, a feature also noted in \cite{Pic2word,SEARLE}. Therefore, we set the Slerp balancing scalar value $\alpha$ to 0.9, 0.8, and 0.8 for CIRR, CIRCO, and FashionIQ, respectively, taking into account the unique features of each dataset.

\subsection{Main Results}
\label{sec:Main Results}

\begin{table}[!t]
\centering
\caption{Retrieval results on \textit{CIRR test set}. The best scores are marked in bold, while the second best are underlined.}
\begin{adjustbox}{width=0.85\textwidth}
\begin{tabular}{l|l|c|c|c|c|c|c|c}
\toprule

 \multirow{2}{*}{Backbone} & \multirow{2}{*}{Method} & \multicolumn{4}{c|}{Recall@K} & \multicolumn{3}{c}{$\text{Recall}_{\text{subset}}$@K} \\ \cmidrule{3-9}
& & K=1 & K=5 & K=10 & K=50 & K=1 & K=2 & K=3 \\ \midrule
 \multirow{6}{*}{\makecell{CLIP-\\ViT-B/32}} & Image-only & ~ 6.99 ~ & ~ 23.23 ~ & ~ 34.31 ~ & ~ 58.89 ~ & ~ 20.46 ~ &  ~ 40.46 ~ & ~ 60.87 ~ \\
 & Text-only & 19.61 &  43.90 & 55.42 & 78.27 & \textbf{62.24} & \textbf{80.99} & \underline{90.36} \\
  & PALAVRA$^\dagger$ \cite{PALAVRA}  & 16.62 & 43.49 & 58.51&  83.95 & 41.61 & 65.30 &  80.94  \\
    & SEARLE$^\dagger$ \cite{SEARLE}  & 23.57 &  \underline{52.80}  & \underline{66.48} & \textbf{89.59}  & 54.80  & 76.70  & 88.10 \\
      & Slerp  & \underline{24.22} & 50.94 & 63.49 & 84.92 & 57.86 & 78.27 & 89.25 \\
   & Slerp + TAT  & \textbf{28.19} & \textbf{55.88} & \textbf{68.77} & \underline{88.51} & \underline{61.13} & \underline{80.63} & \textbf{90.68} \\ \midrule

 \multirow{7}{*}{\makecell{CLIP-\\ViT-L/14}} & Image-only & 7.33 & 23.01 & 33.25 & 56.24 & 20.84 & 41.61 & 61.04 \\
 & Text-only & 20.92 &  43.98 & 55.42 & 76.77 & \underline{60.41} & \underline{79.74} & \underline{90.36} \\
  & Pic2Word$^\ddagger$ \cite{Pic2word} & 23.90 &  51.70 &  65.30 & 87.80 & - & - & - \\
    & SEARLE$^\ddagger$ \cite{SEARLE}  & 24.22 &  52.41 & 66.29  & \underline{88.63} & 53.71 &  74.63 & 87.61 \\
      & LinCIR$^\ddagger$ \cite{LinCIR}  & \underline{25.04} &  \underline{53.25} & \underline{66.68} & - & 57.11 & 77.37 & 88.89 \\
      & Slerp  & 24.43 &  49.93 & 62.29 & 83.45 & 57.71 & 77.59 & 88.80 \\
   & Slerp + TAT & \textbf{30.94} & \textbf{59.40} & \textbf{70.94} & \textbf{89.18} & \textbf{64.70} & \textbf{82.92} & \textbf{92.31} \\ \midrule

    \multirow{4}{*}{\makecell{BLIP-\\ViT-L/16}} & Image-only & 7.18 & 23.30 & 33.40 & 57.49 & 20.92 & 41.45 & 60.58 \\
 & Text-only & 26.48 & 51.45 & 62.10 & 79.66 & \underline{66.27} & \underline{84.53} & \underline{92.31} \\
      & Slerp  & \underline{28.60} & 	\underline{55.37} & \underline{65.66} & \underline{84.05} & 65.16 & 83.90 & 92.05 \\
   & Slerp + TAT & \textbf{33.98} & \textbf{61.74} & \textbf{72.70} & \textbf{88.94} & \textbf{68.55} & \textbf{85.11} & \textbf{93.21} \\

\bottomrule
\end{tabular}
\end{adjustbox}
\label{table:Comparisons_CIRR}
\end{table}

In this section, we compare our method (Slerp and Slerp + TAT) with existing pseudo-word token based ZS-CIR works \cite{PALAVRA,SEARLE,Pic2word,LinCIR}. The symbol $^\dagger$ denotes retrieval results reported in \cite{SEARLE}, $^\ddagger$ denotes results reported in \cite{LinCIR}, and $-$ indicates that results were not reported.

\noindent \textbf{CIRR.} Retrieval results on CIRR dataset are shown in Table \ref{table:Comparisons_CIRR}. Due to the unique property of the CIR triplets in this dataset, where the textual intents of CIR triplets are frequently unrelated as noted in \cite{Pic2word,SEARLE}, text-only retrieval results achieve the highest scores under $\text{Recall}_{\text{subset}}@1, 2$ for the C-B32 backbone. When we compare Slerp with pseudo-word token-based methods, we find that the performance is comparable. The key to this performance is Slerp's ability to adjust the $\alpha$ parameter appropriately, making the retrieval process feasible according to the properties of the retrieval domain. Moreover, except for the C-B32 backbone $\text{Recall}@50$ case (where our method achieves the second-best result), our Slerp + TAT approach significantly outperforms previous works and image/text-only cases for both CLIP and BLIP-based backbones, demonstrating the benefits of both proposals.

\begin{table}[!t]
\centering
\caption{Retrieval results on \textit{CIRCO test set}. The best scores are marked in bold, while the second best are underlined.}
\begin{adjustbox}{width=0.8\textwidth}
\begin{tabular}{l|l|c|c|c|c}
\toprule
Backbone & Method & ~ mAP@5 ~ & ~ mAP@10 ~ & ~ mAP@25 ~ & ~ mAP@50 ~ \\ \midrule
 \multirow{6}{*}{\makecell{CLIP-\\ViT-B/32}} & Image-only & 1.51 & 1.86 & 2.31 & 2.64 \\ 
& Text-only & 2.50 & 2.61 & 2.91 & 3.09 \\ 
& PALAVRA$^\dagger$ \cite{PALAVRA} & 4.61 & 5.32 & 6.33 & 6.80 \\ 
& SEARLE$^\dagger$ \cite{SEARLE} & \underline{8.86} & \underline{9.43} & \underline{10.55} & \underline{11.23} \\ 
& Slerp & 6.35 & 7.11 & 8.12 & 8.75 \\ 
& Slerp + TAT & \textbf{9.34} & \textbf{10.26} & \textbf{11.65} & \textbf{12.33} \\ 
\midrule
 \multirow{7}{*}{\makecell{CLIP-\\ViT-L/14}} & Image-only & 2.50 & 3.09 & 3.93 & 4.42 \\
& Text-only & 3.30 & 3.65 & 4.08 & 4.38 \\ 
& Pic2Word$^\ddagger$ \cite{Pic2word} & 8.72 & 9.51 & 10.64 & 11.29 \\ 
& SEARLE$^\ddagger$ \cite{SEARLE} & 11.68 & 12.73 & 14.33 & 15.12 \\ 
& LinCIR$^\ddagger$ \cite{LinCIR} & \underline{12.59} & \underline{13.58} & \underline{15.00} & \underline{15.85} \\ 
& Slerp & 8.76 & 9.84 & 11.30 & 11.99 \\ 
& Slerp + TAT & \textbf{18.46} & \textbf{19.41} & \textbf{21.43} & \textbf{22.41} \\ \midrule

 \multirow{4}{*}{\makecell{BLIP-\\ViT-L/16}} & Image-only & 1.52 & 1.89 & 2.42 & 2.77 \\
& Text-only & 4.24 & 4.59 & 5.20 & 5.53 \\ 
& Slerp & \underline{9.61} & \underline{10.11} & \underline{11.10} & \underline{11.66} \\ 
& Slerp + TAT & \textbf{17.84} & \textbf{18.44} & \textbf{20.24} & \textbf{21.07} \\ 
\bottomrule
\end{tabular}
\end{adjustbox}
\label{table:Comparisons_CIRCO}
\end{table}

\noindent \textbf{CIRCO.} For CIRCO dataset, retrieval results are shown in Table \ref{table:Comparisons_CIRCO}. Despite the increased challenge posed by CIRCO's larger retrieval gallery size and its greater emphasis on image and relative text (textual intent) in the construction of CIR triplets compared to CIRR, our Slerp + TAT approach still significantly outperforms previous methods and image/text-only cases for both CLIP and BLIP across all mAP score metrics. This performance improvement is even more pronounced when a superior VLP model is used as the backbone, as observed in the results between C-B32 and C-L14. Furthermore, it's noteworthy that simply applying Slerp without any fine-tuning can yield the third-best results with the C-B32 and C-L14 backbones. The only method that outperforms Slerp is SEARLE, which requires training with 3M image-text pairs and 5.5M text samples to achieve its retrieval results.

\noindent \textbf{FashionIQ.} For the fashion domain images from FashionIQ dataset, we present the experimental results in Table \ref{table:Comparisons_FashionIQ}. Similar to the results with the natural image datasets, our Slerp-based retrieval results are comparable to those of pseudo-token based methods in fashion domain either. Moreover, our Slerp + TAT method achieves the best average scores in most cases for both backbone types, with the exception of C-B32 R@10, where our Slerp method shows the best results. These results demonstrate the effectiveness of our approach across diverse datasets and retrieval tasks, further validating the robustness and versatility of our proposed methods.

\begin{table}[!t]
\centering
\caption{Retrieval results on \textit{FashionIQ validation set}. The best scores are marked in bold, while the second best are underlined.}
\begin{adjustbox}{width=0.95\textwidth}
\begin{tabular}{l|l|c|c|c|c|c|c|c|c}
\toprule

 \multirow{2}{*}{Backbone} & \multirow{2}{*}{Method} & \multicolumn{2}{c|}{Dress} & \multicolumn{2}{c|}{Shirt}  & \multicolumn{2}{c|}{Toptee} & \multicolumn{2}{c}{Average} \\ \cmidrule{3-10}
& & R@10 & R@50 &R@10 & R@50 & R@10 & R@50 & R@10 & R@50  \\ \midrule
 \multirow{6}{*}{\makecell{CLIP-\\ ViT-B/32}} & Image-only & ~ 3.87 ~ & ~ 10.81 ~ & ~ 7.46 ~ & ~ 14.03 ~ & ~ 6.22 ~ &  ~ 13.36 ~ & ~ 5.85 ~ & ~ 12.73 ~ \\   
& Text-only & 13.58 & 31.78 & 20.26 & 35.28 & 20.19 & 40.49 & 18.01 & 35.85 \\ 
&  PALAVRA$^\dagger$ \cite{PALAVRA} & 17.25  & 35.94 & 21.49 & 37.05 & 20.55 & 38.76 & 19.76 & 37.25 \\
&  SEARLE$^\dagger$ \cite{SEARLE} & 18.15  & 38.57  & \textbf{24.83} & \underline{41.11} & 25.60 & 46.25 & 22.86 & 41.98 \\
& Slerp & \textbf{20.53} & \underline{41.00}  & \underline{23.75} & 40.92 & \textbf{26.98} & \underline{46.77} & \textbf{23.75} & \underline{42.90} \\
& Slerp + TAT & \underline{19.24} & \textbf{42.14}  & 23.06 & \textbf{41.95} & \underline{26.57} & \textbf{47.78} & \underline{22.96} & \textbf{43.96} \\
\midrule
 \multirow{7}{*}{\makecell{CLIP-\\ ViT-L/14}}  & Image-only & 4.86 & 12.99 & 11.04 & 20.22 & 8.67 & 16.52 & 8.19 & 16.58 \\
& Text-only & 14.33 & 32.57 & 20.46 & 33.61 & 21.72 & 39.32 & 18.84 & 35.17 \\
&  Pic2Word$^\ddagger$ \cite{Pic2word} & 20.00 & 40.20 & 26.20 & 43.60 & 27.90 & 47.40 & 24.70 & 43.70 \\
&  SEARLE$^\ddagger$ \cite{SEARLE} & 20.48 & 43.13  & 26.89 & 45.58 & 29.32 & 49.97 & 25.56 & 46.23 \\
&  LinCIR$^\ddagger$ \cite{LinCIR} & 20.92 & \underline{42.44} & \underline{29.10} & \bf{46.81} & 28.81 & \underline{50.18} & 26.28 & \underline{46.49} \\
& Slerp &  \underline{21.96} & 41.55 & 28.66 & 43.96 & \underline{30.24} & 48.34 & \underline{26.95} & 44.62 \\
& Slerp + TAT &  \textbf{23.35}  & \textbf{45.12} & \textbf{29.64} & \underline{46.47} & \textbf{31.97} & \textbf{51.20} & \textbf{28.32} & \textbf{47.60} \\ \midrule

 \multirow{4}{*}{\makecell{BLIP-\\ ViT-L/16}}  & Image-only &  4.16 & 11.40 & 8.78 & 18.11 & 7.39 & 15.45 & 3.58 & 9.37 \\
& Text-only & 18.05 & 34.90 & 21.39 & 36.36 & 24.73 & 42.94 & 28.16 & 47.93 \\
& Slerp &  \underline{22.91} & \underline{42.39} & \underline{27.33} & \underline{45.25} & \underline{32.33} & \underline{50.48} & \underline{31.89} & \underline{51.60} \\
& Slerp + TAT &  \textbf{29.15}  & \textbf{50.62} & \textbf{32.14} & \textbf{51.62} & \textbf{37.02} & \textbf{57.73} & \textbf{32.77} & \textbf{53.32} \\

\bottomrule
\end{tabular}
\end{adjustbox}
\label{table:Comparisons_FashionIQ}
\end{table}

\subsection{Further Analysis}
\label{sec:Further Analysis}

\begin{figure}[!t]
\centering
  \subcaptionbox{mAP@K
  \label{fig:mAP_plot}}{\includegraphics[height=3cm]{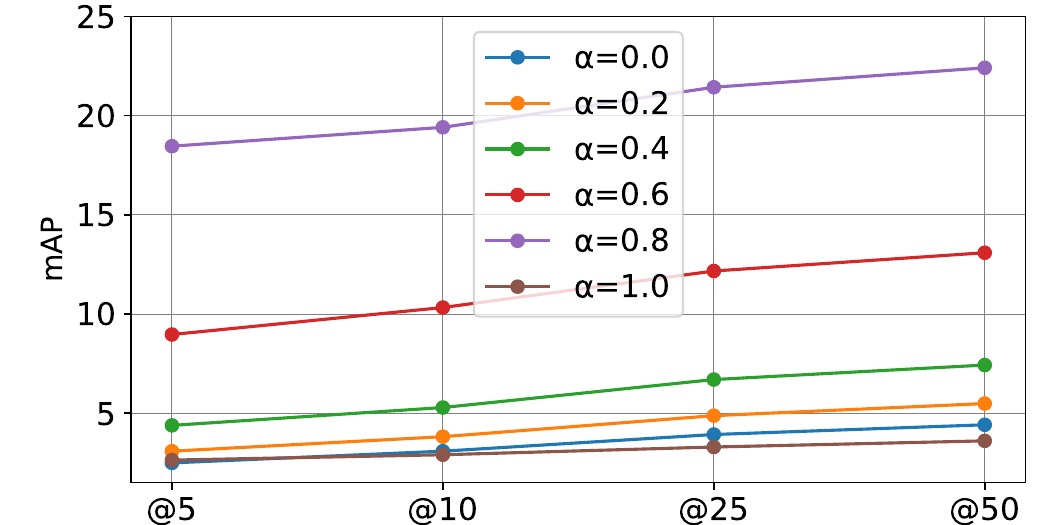}}
  \subcaptionbox{Recall@K
  \label{fig:Recall_plot}}{\includegraphics[height=3cm]{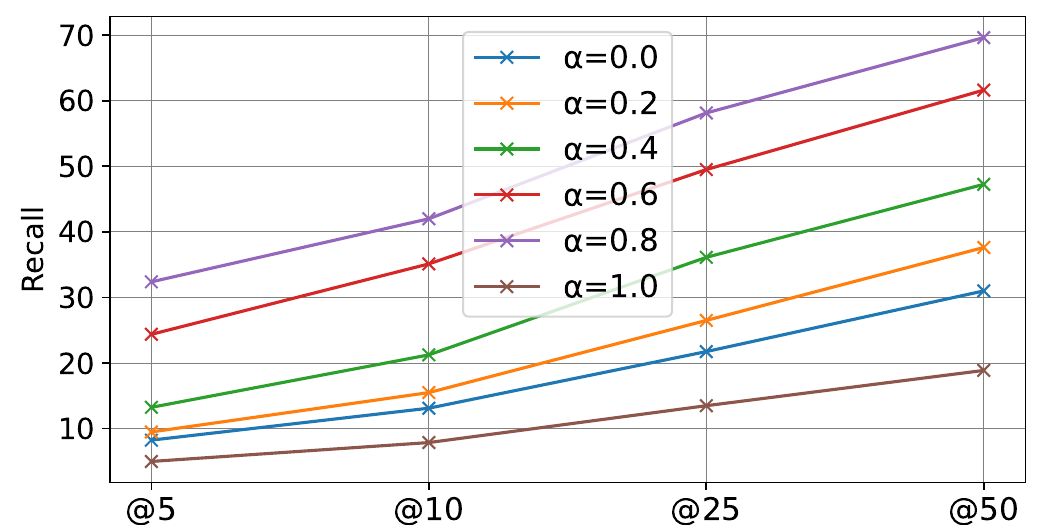}}
\caption{mAP and Recall scores by varying the $\alpha$ of Slerp with CLIP-ViT-L/14 model on \textit{CIRCO test set}.}
\label{fig:Slerp_ablation}
\vspace{-1.5em}
\end{figure}

\noindent \textbf{Ablation Study.} In order to validate our method, we conduct an ablation study on the Slerp and TAT schemes. For Slerp, we vary the balancing value $\alpha$ and displayed the results in Figure \ref{fig:Slerp_ablation}. From these results, we observe that increasing the weight given to text embedding improves both mAP and Recall performance, up to $\alpha=0.8$. However, the image also plays a significant role, as evidenced by the significant drop in performance when the image has no effect ($\alpha=1.0$).

For TAT, we perform experiments with diverse setups and display the results in Table \ref{table:Ablation_CIRCO}. Please refer to Table \ref{table:Comparisons_CIRCO} for a comparison with previous works. We experiment with (a, b, c, d) to evaluate the impact of the training dataset by changing it during TAT. For (a), even though TAT is trained with only 0.58M image-text pairs, which is only 20\% of the CC3M used in \cite{Pic2word,LinCIR}, TAT significantly outperforms previous methods. For (b), even when only a subset of CC3M is utilized for training, our TAT still outperforms previous methods by a large margin. We employ (c, d) to test whether scaling the dataset to a larger scale than our baseline (Laion-2M) would affect the performance, and we find that it does not have much impact. For (e, f), we aim to compare text-anchoring used in TAT with no-anchor: applying LoRA to both the image and text encoder for fine-tuning, and image-anchor: applying LoRA to the text encoder and fixing the image encoder. From the results, we demonstrate that text-anchoring is the most effective training method. For (g), we apply the same Laion-2M dataset to Pic2Word \cite{Pic2word} to verify that the dataset is not the only factor contributing to our performance, and we observe that the performance gain with Laion-2M in Pic2Word is marginal. Additionally, for (h), we apply LoRA to Pic2Word to show that LoRA is not the powerful factor in our proposed design, which also shows a marginal performance gain compared to the original Pic2Word baseline.

\begin{table}[!t]
\centering
\caption{Ablation study results with CLIP-ViT-L/14 backbone on \textit{CIRCO test set}. Except for (g, h), retrieval results are obtained with Slerp + TAT.}
\begin{adjustbox}{width=0.82\textwidth}
\begin{tabular}{l|c|c|c|c}
\toprule
 Ablation & ~ mAP@5 ~ & ~ mAP@10 ~ & ~ mAP@25 ~ & ~ mAP@50 ~ \\ \midrule
(a) LLaVA-Align (0.58M) & 17.05 & 18.23 & 20.11 & 21.05 \\ 
(b) CC3M (2.3M) & 16.98 & 17.82 & 19.62 & 20.58 \\ 
(c) Laion-4M & 18.45 & 19.50 & 21.53 & 22.52 \\ 
(d) Laion-8M & 18.23 & 19.21 & 21.29 & 22.26 \\ 
(e) None-anchoring & 8.26 & 8.90 & 10.07 & 10.71 \\ 
(f) Image-anchoring & 7.54 & 7.73 & 8.79 & 9.30 \\ 
(g) Pic2Word-Laion-2M & 8.93 & 9.96 & 11.50 & 12.02 \\ 
(h) Pic2Word-LoRA & 8.89 & 9.84 & 11.33 & 11.60 \\ 

\bottomrule
\end{tabular}
\end{adjustbox}
\label{table:Ablation_CIRCO}
\end{table}

\begin{table}[!t]
\centering
\caption{Retrieval results with fine-tuned BLIP model on \textit{CIRR test set}.}
\begin{adjustbox}{width=0.72\textwidth}
\begin{tabular}{l|c|c|c|c|c|c|c}
\toprule

\multirow{2}{*}{Method} & \multicolumn{4}{c|}{Recall@K} & \multicolumn{3}{c}{$\text{Recall}_{\text{subset}}$@K} \\ \cmidrule{2-8}
& K=1 & K=5 & K=10 & K=50 & K=1 & K=2 & K=3 \\ \midrule
CASE \cite{CASE} & ~ 35.40 ~ & ~ 65.78 ~ & ~ \textbf{78.53} ~ & ~ \textbf{94.63} ~ & ~ 64.29 ~ &  ~ 82.66 ~ & ~ 91.61 ~ \\
\midrule
CoVR \cite{CoVR} & 38.48 & \textbf{66.70} & 77.25 & 91.47 & 69.28 & 83.76 & 91.11 \\
\midrule
Slerp & \textbf{39.08} & 65.57 & 75.45 & 89.83 & \textbf{72.96} & \textbf{88.10} & \textbf{94.87} \\

\bottomrule
\end{tabular}
\end{adjustbox}
\label{table:Slerp_finetune}
\end{table}

\noindent \textbf{Applying Slerp on Fine-tuned Models.} To further demonstrate the advantage of the Slerp-based ZS-CIR, we conduct an experiment with Slerp on the fine-tuned VLP model (BLIP fine-tuned on COCO \cite{COCO} train set) and show the results in Table \ref{table:Slerp_finetune}. It's important to note that, unlike CASE \cite{CASE} and CoVR \cite{CoVR} which utilize additional visual data samples to improve the BLIP model's performance, our Slerp does not utilize any further training and simply interpolates the original BLIP's image and text embeddings. Nevertheless, Slerp achieves the best results in Recall@1 and all of the $\text{Recall}_{\text{subset}}$ cases, verifying its effectiveness and applicability on existing fine-tuned VLP model for retrieval.

\begin{table}[!t]
\centering
\caption{Supervised trained retrieval results on \textit{CIRR test set}.}
\begin{adjustbox}{width=0.9\textwidth}
\begin{tabular}{l|l|c|c|c|c|c|c|c}
\toprule
 \multirow{2}{*}{Backbone} & \multirow{2}{*}{Pretrained Weight} & \multicolumn{4}{c|}{Recall@K} & \multicolumn{3}{c}{$\text{Recall}_{\text{subset}}$@K} \\ \cmidrule{3-9}
& & K=1 & K=5 & K=10 & K=50 & K=1 & K=2 & K=3 \\ \midrule
 \multirow{2}{*}{\makecell{CLIP-\\ ViT-B/32}} & Original CLIP & ~ 30.99 ~ & ~ 61.28 ~ & ~ 73.69 ~ & ~ 91.40 ~ & ~ 59.64 ~ &  ~ 79.95 ~ & ~ 90.65 ~ \\
  & TAT-trained CLIP & \textbf{33.98} & \textbf{65.06} & \textbf{76.87} & \textbf{92.63} & \textbf{61.78} & \textbf{81.52} & \textbf{91.18} \\
\midrule
 \multirow{2}{*}{\makecell{CLIP-\\ ViT-L/14}} & Original CLIP & 30.31 & 61.11 & 73.52 & 90.75 & 60.84 & 80.22 & 90.02 \\
  & TAT-trained CLIP & \textbf{36.58} & \textbf{67.71} & \textbf{78.19} & \textbf{92.72} & \textbf{65.52} & \textbf{84.07} & \textbf{92.89} \\
\bottomrule
\end{tabular}
\end{adjustbox}
\label{table:Supervised_CIRR}
\end{table}

\begin{table}[!t]
\centering
\caption{Supervised trained retrieval results on \textit{FashionIQ validation set}.}
\begin{adjustbox}{width=0.95\textwidth}
\begin{tabular}{l|l|c|c|c|c|c|c|c|c}
\toprule

 \multirow{2}{*}{Backbone} & \multirow{2}{*}{Pretrained Weight} & \multicolumn{2}{c|}{Dress} & \multicolumn{2}{c|}{Shirt}  & \multicolumn{2}{c|}{Toptee} & \multicolumn{2}{c}{Average} \\ \cmidrule{3-10}
& & R@10 & R@50 &R@10 & R@50 & R@10 & R@50 & R@10 & R@50  \\ \midrule
 \multirow{2}{*}{\makecell{CLIP-\\ ViT-B/32}} & Original CLIP & ~ 25.08 ~ & ~ 48.79 ~ & ~ 32.68 ~ & ~ 52.80 ~ & ~ 33.35 ~ &  ~ 56.09 ~ & ~ 30.37 ~ & ~ 52.56 ~ \\   
& TAT-trained CLIP & \textbf{25.48} & \textbf{49.98} & \textbf{33.27} & \textbf{53.29} & \textbf{34.88} & \textbf{57.22} & \textbf{31.21} & \textbf{53.50} \\ 
\midrule
 \multirow{2}{*}{\makecell{CLIP-\\ ViT-L/14}}  & Original CLIP & 29.65 & 52.80 & 39.01 & 57.85 & 38.25 & 59.41 & 35.64 & 56.69 \\
& TAT-trained CLIP & \textbf{30.09} & \textbf{55.73} & \textbf{39.55} & \textbf{58.93} & \textbf{40.44} & \textbf{62.47} & \textbf{36.69} & \textbf{59.04} \\

\bottomrule
\end{tabular}
\end{adjustbox}
\label{table:Supervised_FashionIQ}
\end{table}

\noindent \textbf{Improving Supervised CIR: Replacing VLP with TAT-trained Model.} To further assess the utilization of the TAT-tuned VLP model, we apply it to supervised CIR tasks by replacing the original CLIP model in Combiner \cite{Combiner} training pipeline with our TAT-tuned CLIP. From the results listed in Tables \ref{table:Supervised_CIRR} and \ref{table:Supervised_FashionIQ}, which cover both natural and fashion image domains, we observe a significant improvement in retrieval performance. This confirms that TAT learning of the VLP model not only improves zero-shot performance in the composed image retrieval task, but also serves as an effective initial checkpoint for supervised composed image retrieval scenarios.

\begin{figure}[!t]
\centering
  \subcaptionbox{Top-5 retrieval results on different $\alpha$ value for Slerp.
  \label{fig:qualitative_a}}{\includegraphics[width=0.99\linewidth]{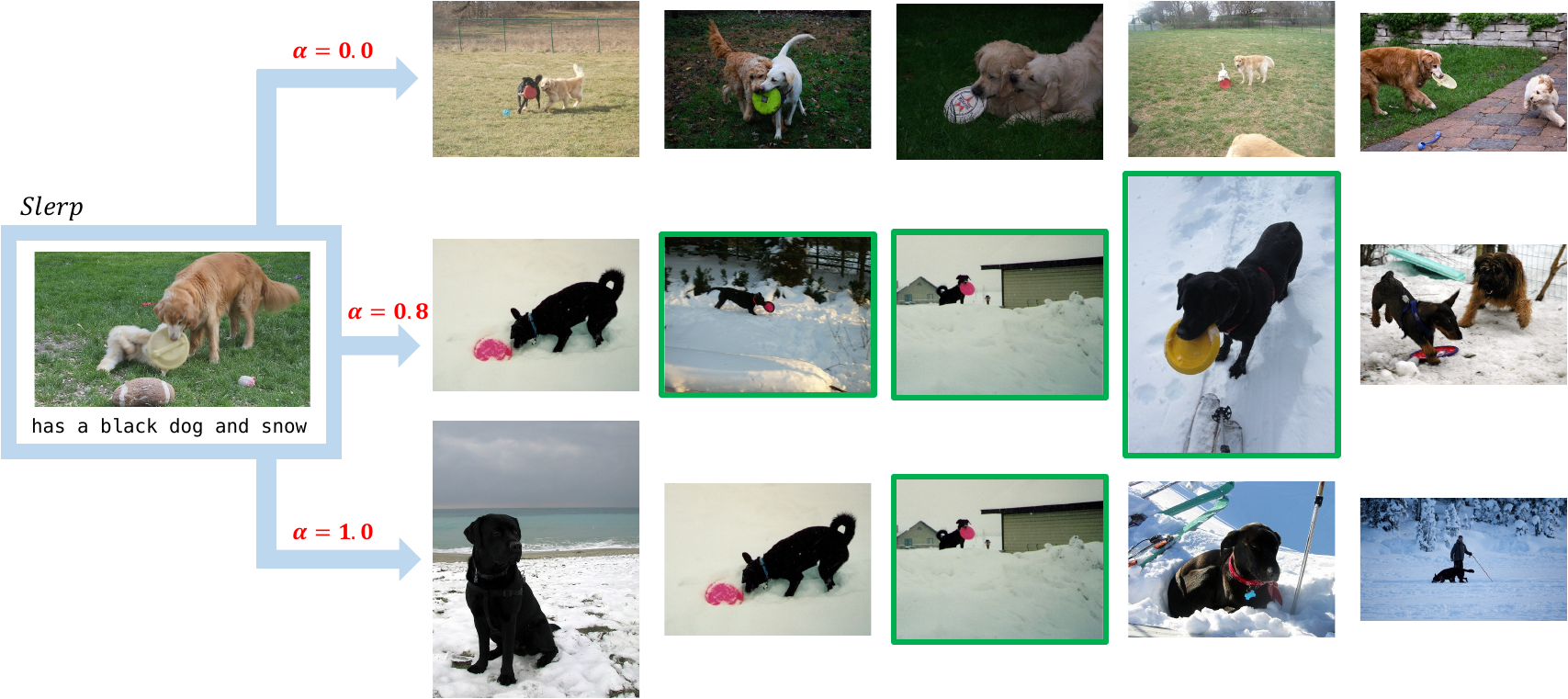}}
  \subcaptionbox{Top-1 retrieval results on different $\alpha$ value for Slerp.
  \label{fig:qualitative_b}}{\includegraphics[width=0.99\linewidth]{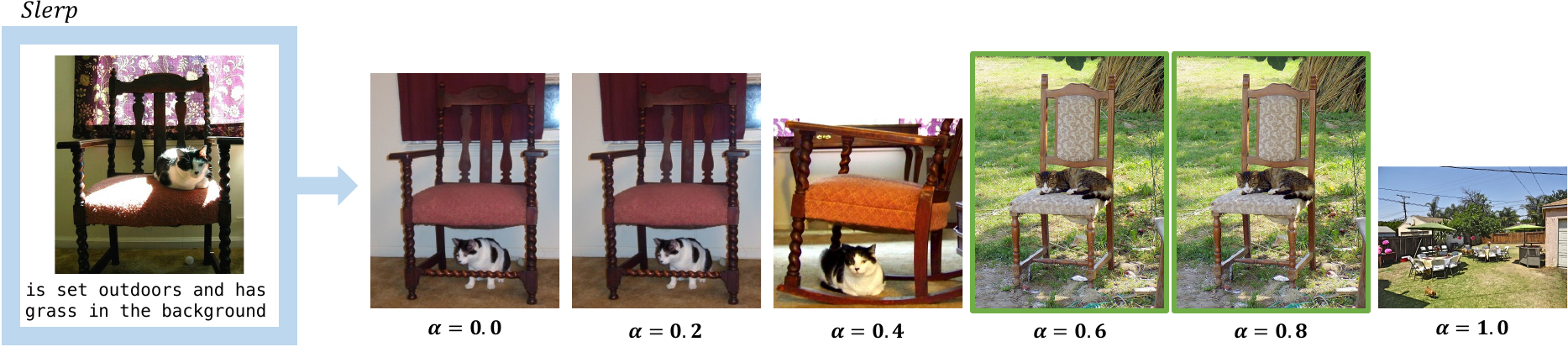}}
\caption{Qualitative results on \textit{CIRCO validation set}. Green box denotes ground truth.}
\label{fig:Qualitative}
\end{figure}

\noindent \textbf{Qualitative Results.} Figure \ref{fig:Qualitative} shows retrieval results using the Slerp + TAT with C-L14 model. From the results, we can confirm that $\alpha$ successfully balances the contribution between the image and text query, enabling user-side adaptation. Moreover, retrieved images are highly relevant, demonstrating the zero-shot capabilities of our proposed method.

\section{Discussion}
\label{sec:Discussion}

\noindent \textbf{Potential Impact.} The proposed Slerp-based ZS-CIR method offers a simple and effective solution to the challenge of establishing a compositional understanding between image and text, without requiring additional training. This could potentially streamline the image retrieval process, making it more efficient and precise. The TAT strategy enhances the performance of the VLP model in ZS-CIR by aligning the distribution of image embeddings with that of text embeddings. This could also improve the performance of supervised CIR by providing better initial checkpoints. Moreover, TAT proves to be resource-efficient as it can deliver decent performance even when trained with far fewer image-text training pairs and requires only a single epoch of training.

\noindent \textbf{Limitation.} While ZS-CIR methods, including our Slerp-based approach, can effectively integrate image and text, they have not yet been demonstrated for different types of composed retrieval. For instance, scenarios where the query consists of both image and text and the retrieval gallery is built with text, or where the query and the retrieval gallery are both composed of image and text samples. This necessitates the consideration of more benchmarks and leaves an open question for the retrieval community to further explore.

\section{Conclusion}
\label{sec:Conclusion}
In this paper, we introduce a Slerp-based zero-shot composed image retrieval method and a text-anchored-tuning strategy. Both present significant advancements in the field of vision-language compositional retrieval tasks. The Slerp-based method provides a simple yet effective solution for integrating image and text, achieving performance on par with existing methods without the need for additional training. The text-anchored-tuning strategy successfully redistributes image samples to closely align with the corresponding text samples in the embedding space, thereby enhancing the performance of the Slerp-based search. Extensive experimental results show that our approach not only achieves superior training efficiency but also demonstrates broader applicability with the state-of-the-art zero-shot composed image retrieval performance.

\bibliographystyle{splncs04}
\bibliography{main}

\appendix

\section{More Qualitative Results}

\begin{figure}[!h]
\centering
\includegraphics[width=0.99\linewidth]{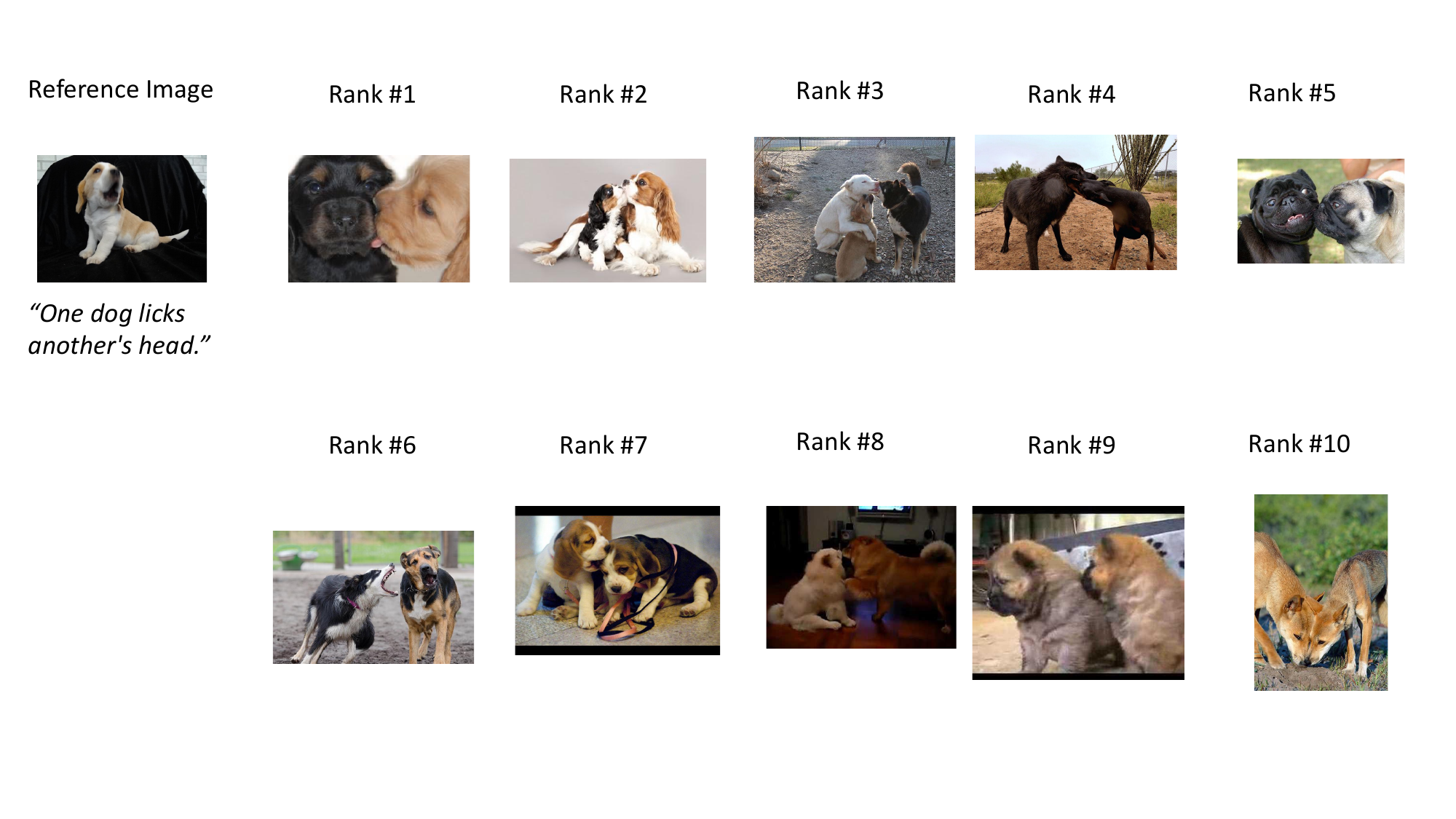}
\includegraphics[width=0.99\linewidth]{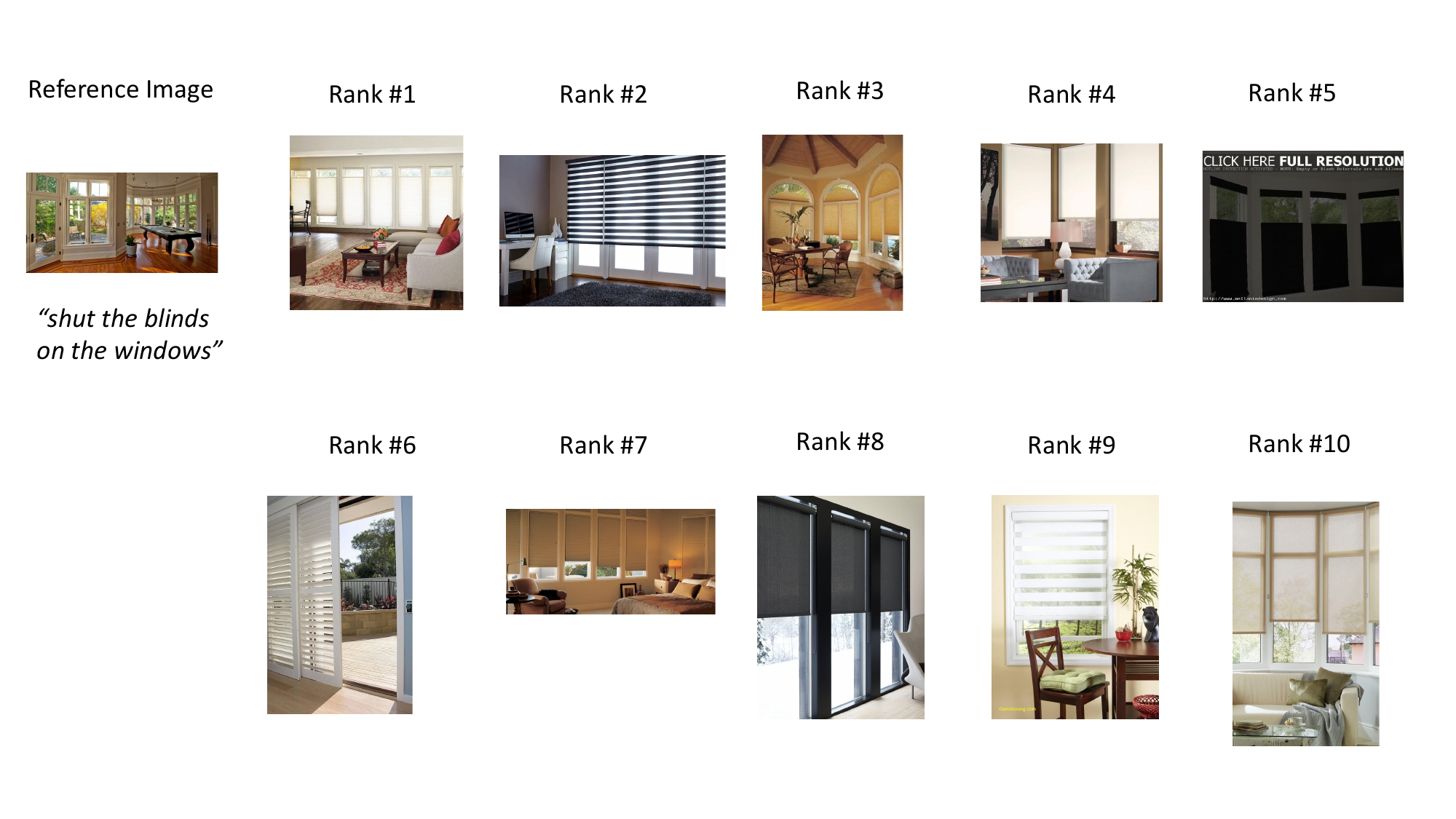}
\caption{Retrieval results on \textit{CIRR test set}.}
\label{fig:supp_cirr}
\end{figure}

\begin{figure}[!h]
\centering
\includegraphics[width=0.99\linewidth]{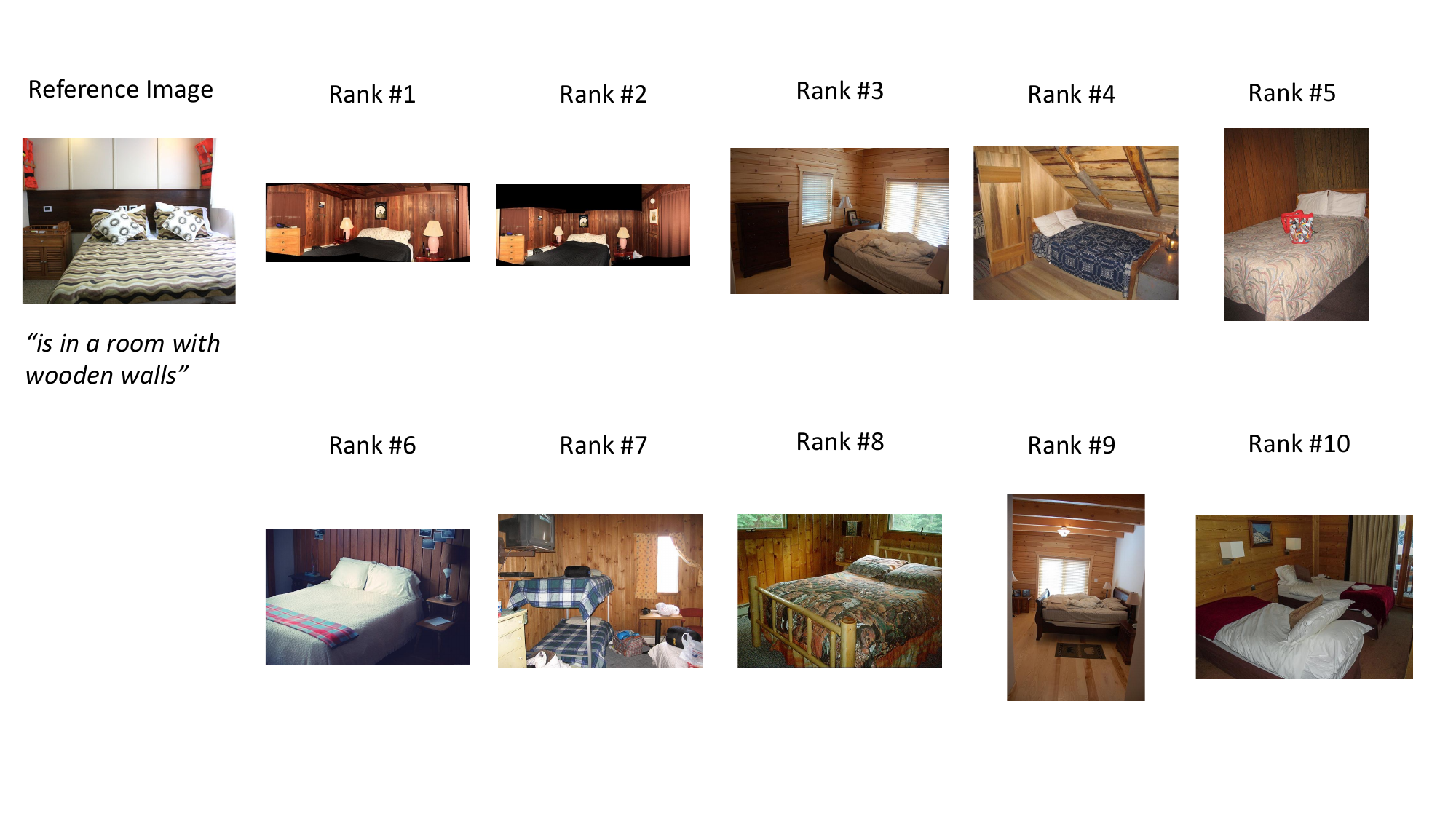}
\includegraphics[width=0.99\linewidth]{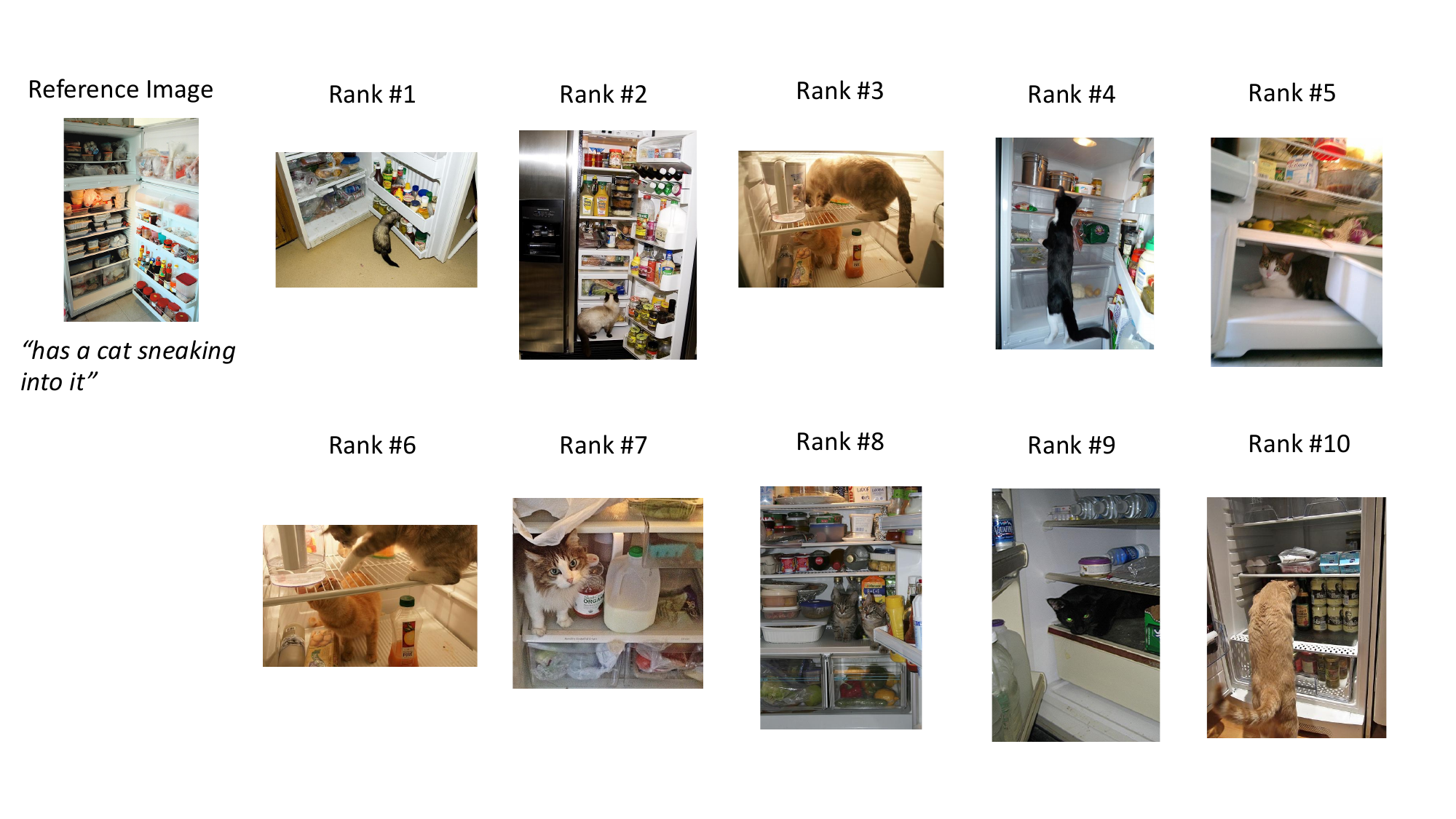}
\caption{Retrieval results on \textit{CIRCO test set}.}
\label{fig:supp_circo}
\end{figure}

\begin{figure}[!h]
\centering
\includegraphics[width=0.99\linewidth]{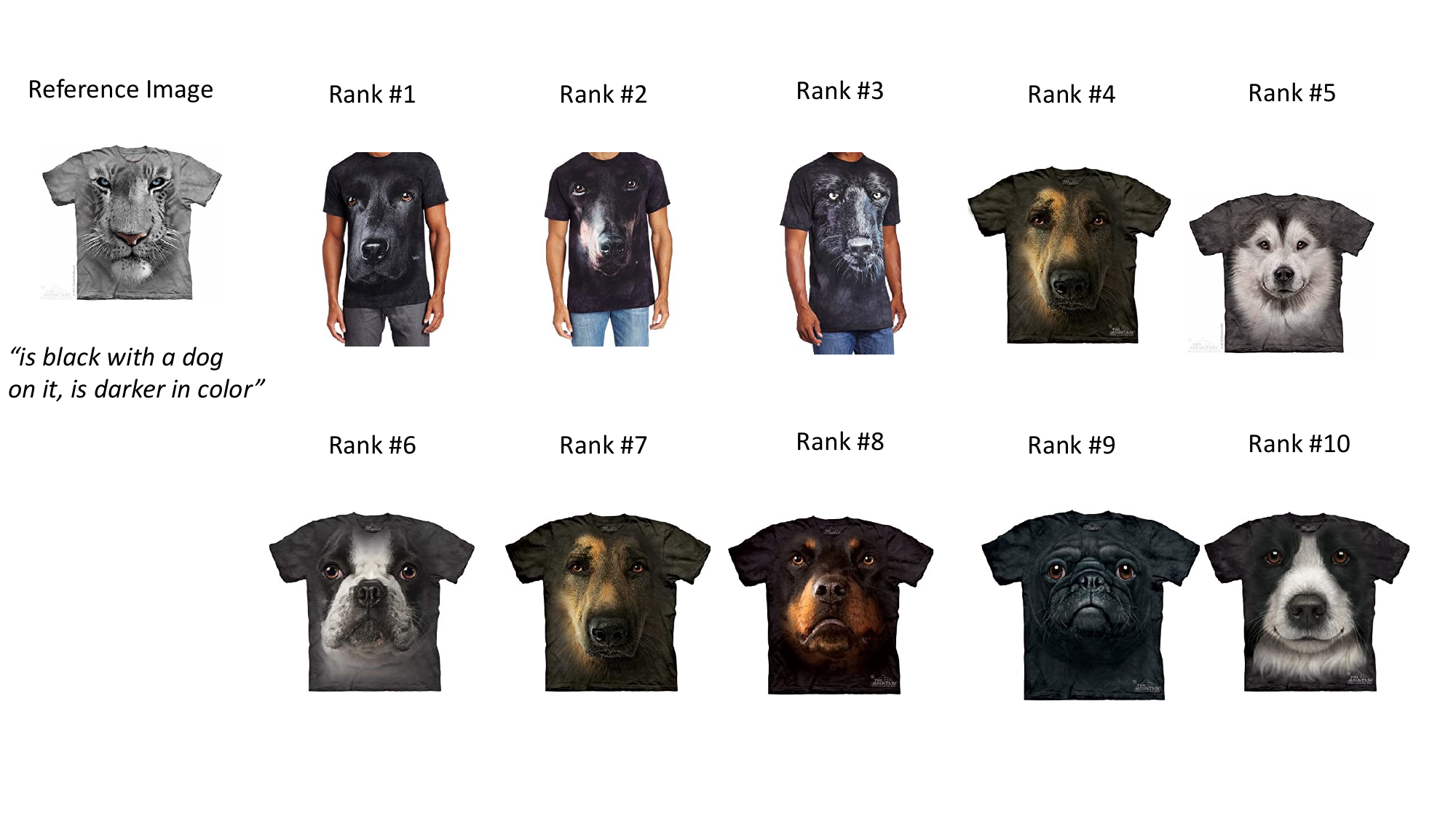}
\includegraphics[width=0.99\linewidth]{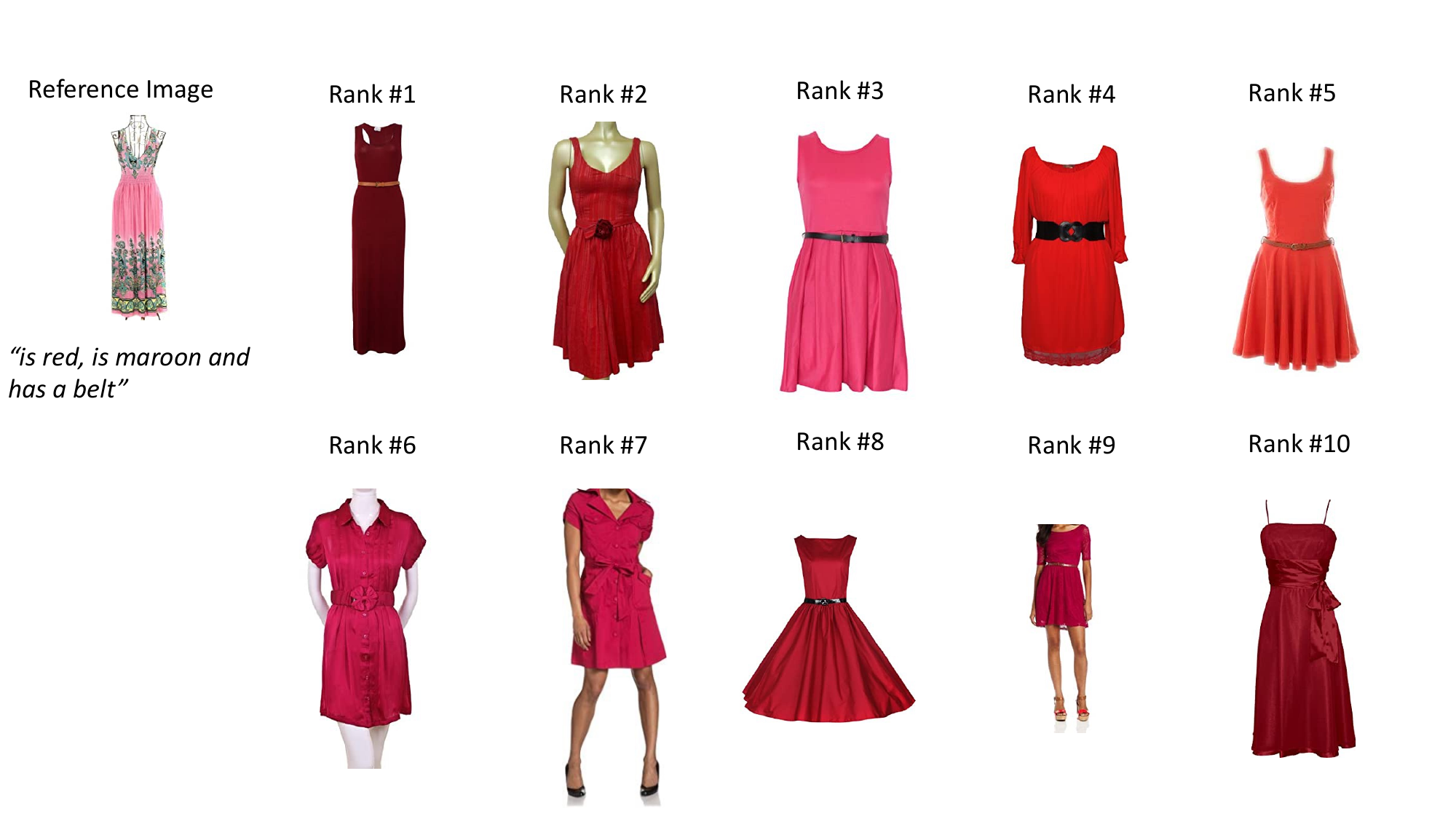}
\caption{Retrieval results on \textit{FashionIQ validation set}.}
\label{fig:supp_fashionIQ}
\end{figure}

In this supplementary material, we provide additional qualitative results for our Slerp + TAT with BLIP-ViT-L/16 \cite{BLIP} backbone on CIR benchmarks. Refer to Figure \ref{fig:supp_cirr} for CIRR \cite{CIRR}, Figure \ref{fig:supp_circo} for CIRCO \cite{SEARLE}, and Figure \ref{fig:supp_fashionIQ} for FashionIQ \cite{fashionIQ}.

\end{document}